%% file: main.tex
\documentclass[letterpaper]{article}
\usepackage{aaai2027}
\usepackage{times}
\nocopyright

\usepackage{helvet}
\usepackage{courier}
\usepackage{graphicx} 
\usepackage{natbib}
\usepackage{xspace}
\usepackage{amsfonts}
\usepackage{amsmath}
\usepackage{multirow}
\usepackage{booktabs}
\usepackage[table]{xcolor}
\usepackage{siunitx}
\usepackage{pifont}
\newcommand{\cmark}{\ding{51}}%
\newcommand{\xmark}{\ding{55}}%

\def\OURS{{INTERACT}\xspace}
\begin{document}
%
\title{INTERACT: Learning Interaction-Aware World Dynamics from Driving Videos for Autonomous Driving}
\author{
    Zhan Chen\textsuperscript{\rm 1},
    Songyan Zhang\textsuperscript{\rm 1},
    Wenhui Huang\textsuperscript{\rm 1},
    Zirui Li\textsuperscript{\rm 1},\\
    Zhidong Wang\textsuperscript{\rm 1},
    Qihang Huang\textsuperscript{\rm 1},
    Shiyu Fang\textsuperscript{\rm 2},
    Chen Lv\textsuperscript{\rm 1}\protect\thanks{Corresponding author.}
}

\affiliations{
    \textsuperscript{\rm 1}Nanyang Technological University, Singapore\\
    \textsuperscript{\rm 2}Tongji University, Shanghai, China
}
\maketitle
\begin{abstract}
\begin{quote}

Safe autonomous driving requires more than recognizing the current scene: the ego vehicle must anticipate how surrounding traffic participants will evolve under its own motion. However, existing end-to-end driving models typically encode scenes into dense visual or BEV features, leaving object-level interactions implicit. We propose INTERACT (\textbf{INTER}action-aware \textbf{A}ction-\textbf{C}onditioned \textbf{T}rajectory learning), a three-stage framework that learns a slot-based object-centric world model from unlabeled driving videos and transfers it to trajectory planning. INTERACT first learns recurrent slots that group dense visual features into temporally structured scene-entity representations without object annotations. It then infers latent ego actions from slot context and video transitions, and uses them to condition future slot prediction, capturing how ego motion and surrounding entities co-evolve. Finally, the learned slots, latent actions, and future slot predictions are distilled into a trajectory planner to guide trajectory generation and selection. Extensive experiments on NAVSIM and nuScenes validate the effectiveness of INTERACT. In particular, it achieves state-of-the-art performance on NAVSIM, with improvements in safety-critical collision metrics.

\end{quote}
\end{abstract}

\input{intro}

\input{related}

\input{method}

\input{exp}

\input{conclu}

\bibliography{ref}
\input{appen}
\end{document}

%% file: intro.tex
\section{Introduction}
\noindent Autonomous driving is inherently interactive: planning a safe trajectory requires anticipating how surrounding traffic participants may evolve under ego motion, and how their future states may constrain ego planning. Classical prediction-and-planning pipelines model such interactions through explicit object-level interfaces, where agents are detected and tracked, their future motions are forecast, and the resulting structured states are passed to the planner~\cite{e2e}. While effective, such pipelines depend on costly object-level annotations, including detection labels, instance masks, tracking annotations, and future trajectories of non-ego agents. Recent end-to-end (E2E) driving models reduce this reliance on hand-designed modules by learning directly from sensor observations and ego trajectories~\cite{autovla,wisead,stp3,emma}. However, they often represent scenes with dense visual, BEV, or query features, where traffic participants, road structures, background appearance, and ego-induced visual changes are entangled. As a result, both object-level interactions and ego-action structure remain implicit: the learned representation may capture scene changes, but does not explicitly reveal which entities interact with the ego vehicle, which changes are induced by ego behavior, or how future object states should affect planning.

Raw driving videos provide important cues for recovering these structures without object-level annotations. Across frames, temporal persistence and coherent motion offer natural signals for grouping dense visual features into object-like scene entities. Meanwhile, frame-to-frame visual changes encode the coupled evolution between ego behavior and the surrounding environment, revealing how the scene changes as the ego vehicle moves. Recent progress in slot-based object-centric learning has shown that such entity-level structure can emerge from visual observations through self-supervised learning~\cite{slot,savi,videosaur}. In parallel, video-based latent-action learning suggests that compact action and motion representations can be inferred from unlabeled visual transitions~\cite{lapa,moto,apv,genie,vjepa2}. These two lines of work indicate that unlabeled videos contain both object-centric cues and ego-action cues that are useful for learning structured representations of driving dynamics.

However, extracting object-centric states or latent actions from observed transitions alone is not sufficient for planning. A planner also needs explicit predictions of how the scene will evolve in the future under ego behavior. World models \cite{nwm} address this by learning predictive representations of future scene states, and recent world models for autonomous driving have shown the benefit of predicting future images \cite{drivevlaw0}, occupancy states \cite{occworld}, scene geometry \cite{geoworldad, eponav2}, or latent planning representations~\cite{law}. Nevertheless, most existing driving world models still encode object information, interactions, and ego-induced dynamics into dense or global representations. They neither explicitly model object-centric evolution nor tightly associate future state prediction with ego actions. This leaves a key challenge: learning an object-centric, action-conditioned world model from unlabeled driving videos that jointly represents structured scene entities, latent ego actions, and their coupled future evolution for interaction-aware trajectory planning.

In this work, we propose \OURS, a three-stage framework that learns a compositional object-centric world model of latent scene entities and ego-conditioned dynamics from unlabeled driving videos, and transfers the learned structure to trajectory planning. The first two stages are trained in a fully self-supervised manner. Specifically, Stage I organizes dense visual observations into recurrent slots representing persistent scene entities, while Stage II integrates visual transition cues with historical slot context to infer latent ego actions, and uses these actions to forecast future slot states. Stage III distills the learned dynamics into an E2E planner that jointly predicts trajectory hypotheses, latent actions, and future slots under ego-trajectory supervision and world-model alignment. Our contributions are three-fold:

\begin{enumerate}
    \item \textbf{Object-centric representation from unlabeled driving videos.} 
    We introduce a recurrent slot-based representation that learns temporally persistent latent scene entities from unlabeled driving videos, providing a structured scene-entity interface for trajectory planning.

    \item \textbf{Ego-action-conditioned object-centric world modeling.} 
    We model both how ego behavior can be inferred from visual dynamics and how object slots evolve under ego latent actions, shifting from reconstruction-driven learning to interaction-aware dynamics learning.

    \item \textbf{World-model-guided trajectory planning.} 
    We integrate learned slots, latent actions, and predicted future slots into trajectory planning, providing interaction-aware and action-conditioned cues for trajectory generation and selection. Our method achieves state-of-the-art performance on the NAVSIM v1 and v2 benchmarks, with notable improvements in safety-critical collision metrics.
\end{enumerate}

%% file: related.tex
\section{Related Works}

\subsection{End-to-end Planning.}
Autonomous driving has shifted from modular pipelines to E2E learning-based systems~\cite{e2e}. Classical pipelines decompose driving into perception, prediction, and planning, using detected agents, tracked histories, and map elements as explicit interfaces for interaction reasoning, but often require dense object-level supervision and hand-designed task interfaces. Recent E2E planners instead learn task-relevant representations directly from sensor observations and ego trajectories. Existing methods have explored camera-LiDAR fusion and BEV-based imitation learning~\cite{transfuser}, unified perception-prediction-planning architectures~\cite{uniad,paradrive}, multimodal trajectory generation and selection with diffusion, proposal-centric decoding, distillation, or learned scoring~\cite{diffusiondrivev2,ipad,drivesuprim,drivor}, as well as VLA-based driving models with language and reasoning abilities~\cite{autovla,opendrivevla,omnidrive}. Despite these advances, most planners still operate on dense visual, BEV, or query features, where traffic participants, road structures, and ego-induced changes are entangled. In contrast, \OURS learns an object-centric ego-object world model from raw driving videos and transfers the learned structured dynamics to trajectory planning.

\subsection{World Models for Autonomous Driving.}
World models learn predictive representations of environment dynamics, which are essential for planning under uncertainty. In autonomous driving, recent works explore future scene prediction in the form of generated driving videos, future visual features, occupancy states, or latent planning representations~\cite{law,world4drive,eponav2,drivewm,occworld,drivelaw,geoworldad}, showing that future observations or scene-state prediction can improve downstream planning. However, many of them represent the scene as a holistic latent state or dense feature field, making it difficult to expose object-level interactions and the causal effect of ego behavior on future scene evolution. In parallel, joint-embedding predictive architectures provide a latent predictive paradigm by predicting future representations instead of reconstructing pixels~\cite{vjepa2,nwm}, encouraging semantic and dynamical abstraction while avoiding unnecessary low-level appearance details. Recent driving work has further connected JEPA-style video representations with trajectory planning~\cite{drivejepa}. Building on this latent predictive perspective, \OURS learns a structured, action-conditioned world model for planning, moving beyond holistic future feature prediction toward ego-aware scene dynamics.

\subsection{Video-Based Action and Object-Centric Learning.}
Recent studies show that videos can provide supervision for action and dynamics learning without explicit action labels. Action-free pretraining learns transferable visual representations for control~\cite{apv}. LAPA infers discrete latent actions from frame transitions to pretrain VLA models~\cite{lapa}, while Moto represents robot videos as latent motion tokens for manipulation~\cite{moto}. Generative interactive environments and latent action-conditioned video models further suggest that unlabeled videos contain reusable priors for prediction and control~\cite{genie,vjepa2}. These works show that video transitions reveal latent causes of change, such as actions, motions, or controllable dynamics. However, they mainly focus on manipulation or general video domains, leaving the ego-environment interaction structure in driving videos underexplored.

Object-centric learning offers a complementary way to model structured dynamics by decomposing visual observations into latent entities. Slot Attention learns object-like representations through iterative attention~\cite{slot}, while SAVi introduces temporal recurrence for consistent object binding in videos~\cite{savi}. VideoSAUR further scales object-centric learning to real-world videos using temporal feature similarity as a self-supervised motion cue~\cite{videosaur}. These methods show that object-centric slots can emerge without dense annotations. \OURS introduces slots as the intermediate interface between dense video representations and trajectory planning, and further couples these slots with latent ego actions to model interaction-aware scene evolution.

\begin{figure*}[h]\centering\includegraphics[width=1.00\textwidth]{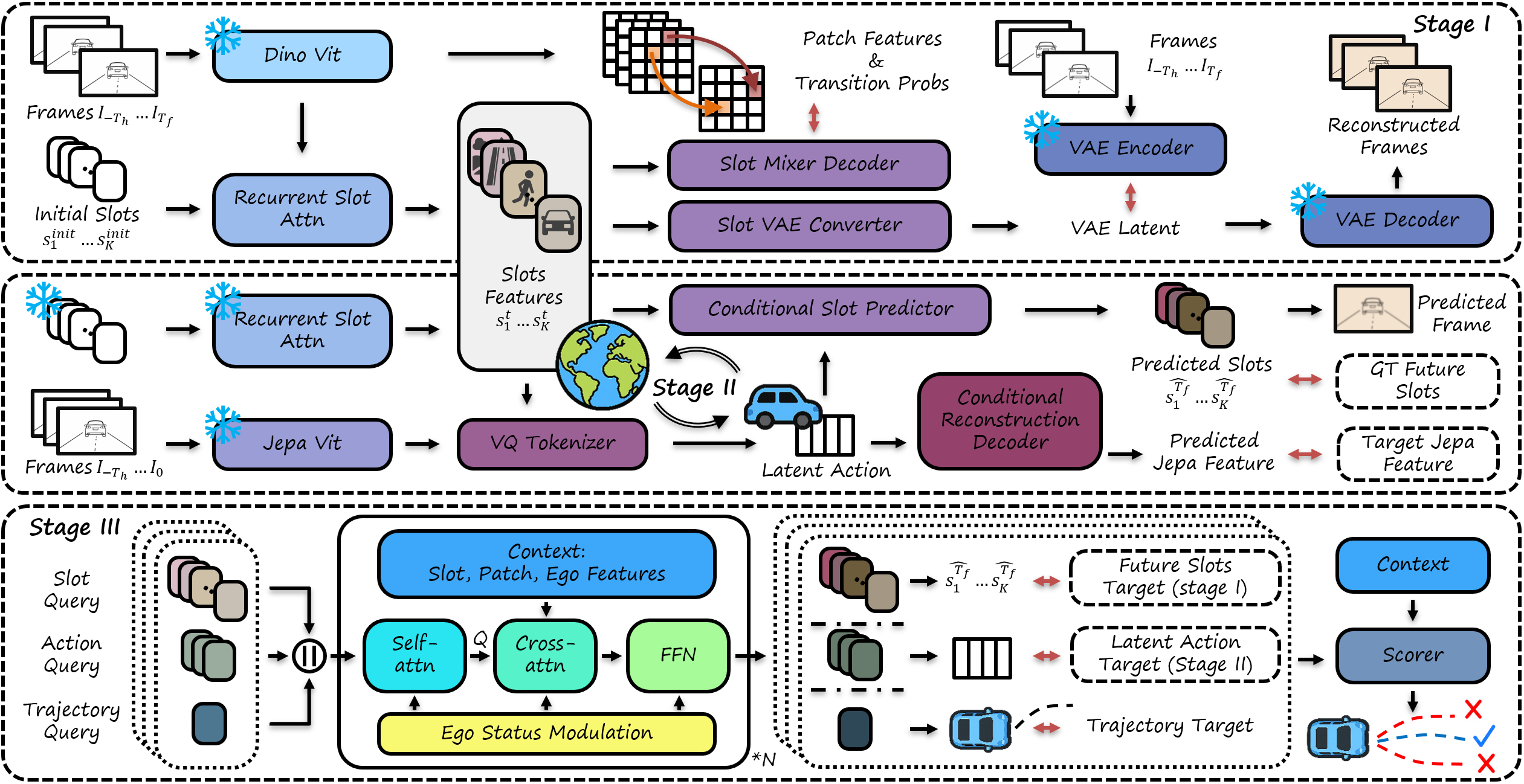}\caption{Overview of the framework. Stage I learns object-centric slot representations. Stage II learns latent ego actions and ego-conditioned future slot dynamics. Stage III distills the learned world dynamics into a trajectory planner, with future-slot and action-token predictions supervised by pseudo-labels from the frozen slot encoder and action tokenizer, respectively.}\label{fig:pipeline}
\end{figure*}

%% file: method.tex
\section{Method}
\subsection{Overview}

\OURS is a three-stage framework that learns an object-centric ego-object world model from raw driving videos and transfers the learned dynamics to trajectory planning. The overall pipeline is illustrated in Fig.~\ref{fig:pipeline}. At a high level, \OURS factorizes driving video dynamics into three complementary components: object-centric scene representation, latent ego-action modeling, and action-conditioned future scene evolution. The model is optimized through a sequential three-stage training procedure. Stage I learns recurrent object-centric slots from frozen visual features, providing a structured latent scene state without object-level supervision. Building on this representation, Stage II learns ego-conditioned dynamics by inferring latent ego actions from visual transitions and historical slots, and predicting future slot evolution under the inferred action. Together, Stages I and II constitute the self-supervised object-centric ego-object world model. Stage III transfers the learned world model to a supervised planner, which jointly predicts scene-conditioned ego actions, future slots, and trajectory hypotheses for interaction-aware planning.


\subsection{Recurrent Object-Centric Video Representation}


Explicit  prediction-and-planning pipelines model interactions over agent histories and map elements. Stage I aims to construct an object-level latent interface from raw videos by learning a recurrent set of slots as latent states of scene elements. These slots are optimized to preserve the spatial, dynamic, and appearance cues required for later world modeling. Specifically, given a video clip $\mathcal{I}_{-T_h:T_f}=\{I_{-T_h},\dots,I_{T_f}\}$, we first extract frozen DINOv2 features
$X_t = E_{\mathrm{dino}}(I_t) \in \mathbb{R}^{P \times D}$,
where $P$ and $D$ denote the number of image patches and the feature dimension, respectively. Following recurrent object-centric video models such as SAVi~\cite{savi}, we bind patch features into $K$ temporally persistent slot representations with a recurrently initialized Slot Attention module~\cite{slot}:

\begin{equation}
    S_t, M_t = \mathrm{SlotAttn}(X_t; S_{t-1}),
    \label{eq:rsa}
\end{equation}
where $S_t=\{{s_t^1,\ldots,s_t^K}\}$ denotes the updated slot set and $M_t$ represents the soft assignment masks between patches and slots. The slots are initialized from a learned Gaussian distribution at the first frame and recurrently initialized from the previous time step afterwards, encouraging temporal consistency and stable object binding across time. 
We supervise the slots with complementary objectives at the feature, temporal, and appearance levels.
To preserve semantic and spatial information, a SlotMixer decoder reconstructs DINO features from the slots, where $M_t$ specifies the soft ownership of each patch by different slots. 
To make the slots sensitive to temporal scene dynamics, we follow VideoSAUR~\cite{videosaur} and train the decoder to predict temporal patch-transition logits $\hat A_t$. 
The target transition distribution is constructed from cosine similarities between the normalized frozen feature of a source patch $p$ and that of a target patch $q$ in the adjacent frame:
\begin{equation}
    T_{t,t+1}(p,q)
    =
    \mathrm{softmax}_{q}
    \left(
    \frac{
    \langle \bar X_t(p), \bar X_{t+1}(q) \rangle
    }{\tau}
    \right).
    \label{eq:patch-transition-logits}
\end{equation}
where the temperature parameter $\tau$ controls the sharpness of the transition distribution. The temporal prediction loss $\mathcal{L}_{\mathrm{temp}}$ is computed as the cross-entropy between $T_{t,t+1}$ and $\mathrm{softmax}(\hat A_t)$.
This biases slots toward grouping patches with consistent semantic and temporal correspondence. Finally, a slot-to-VAE converter maps the spatially composed slot representation to the latent space of a frozen image VAE, providing an efficient appearance-level target without training an image decoder from scratch. The reconstructed DINO features and VAE latents are supervised with mean-squared error losses.
The overall Stage I objective is

\begin{equation}
    \mathcal{L}_{\mathrm{I}}
    =
    \lambda_{\mathrm{feat}}\mathcal{L}_{\mathrm{feat}}
    +
    \lambda_{\mathrm{temp}}\mathcal{L}_{\mathrm{temp}}
    +
    \lambda_{\mathrm{vae}}\mathcal{L}_{\mathrm{vae}}.
    \label{eq:stage1-loss}
\end{equation}
This objective yields compact, temporally stable, and dynamically informative slots, which serve as the object-centric scene state for subsequent ego-conditioned world modeling.

\subsection{Bidirectional Ego-Object World Modeling}

Stage II constructs an object-centric world model that couples latent ego-action inference with action-conditioned prediction of future scene states. An ego action should account for the ego-induced changes in visual observations, while also providing an informative condition for predicting future scene evolution.  
We therefore jointly learn latent ego actions and ego-conditioned future slot dynamics.

Given a training video clip $\mathcal{I}_{-T_h:T_f}$, the frozen Stage I encoder produces historical slots $S_{-T_h:0}$ as the object-level scene interface, and target slots $S_{T_f}^{\star}$ as supervision for future slot prediction. 
In parallel, we use a frozen V-JEPA2 encoder~\cite{vjepa2} to extract motion-aware visual tokens, including context tokens $H^{ctx} = E_{\mathrm{jepa}}(I_{-T_h:0})$ from the history and target tokens $H^{tar} = \operatorname{Tail}(E_{\mathrm{jepa}}(I_{-T_h:T_f}))$ from the full clip, which provide transition cues beyond static semantic features for inferring ego behavior in dynamic driving scenes. Importantly, future observations are used only during Stage II training; the downstream planner observes historical inputs only at inference time. An MFormer then models the visual transition, followed by cross-attention to the historical slots to inject object-centric context into the resulting motion representation:
\begin{equation}
    \widetilde U =
    \mathrm{CrossAttn}
    \bigl(
    \mathrm{MFormer}(H^{ctx}, H^{tar}),\,
    S_{-T_h:0}
    \bigr).
    \label{eq:motion-slot-context}
\end{equation}
Details of the MFormer implementation are provided in Appendix A.
This design allows the action representation to capture frame-level visual transitions together with the scene dynamics encoded by the slots. We then quantize $\widetilde U$ into ego latent action tokens using a tokenizer based on vector quantization (VQ), imposing a compact behavioral bottleneck that encourages recurring visual transitions to share a reusable vocabulary of ego behaviors. 
To improve the expressiveness of the discrete action space, we adopt residual VQ with $L$ codebook levels and obtain the final ego latent action as
\begin{equation}
    z^{ego} = Q_{\psi}(\widetilde U) = \sum_{\ell=1}^{L} q^{(\ell)} ,
    \label{eq:ego-action-token}
\end{equation}
where $q^{(\ell)}$ is the code vector selected at the $\ell$-th residual quantization level. The resulting ego latent action is required to reconstruct the target JEPA feature $\hat H^{tar}$ together with the history context using a transformer decoder.

To further enforce interaction-aware action learning, we train a conditional slot predictor to forecast future slot states based on the inferred ego action. 
Given the last historical slots $S_0$, the predictor first models slot-wise dependencies and then modulates the transition with the ego action:
\begin{equation}
\begin{aligned}
\widetilde S_0 &= S_0 + \mathrm{MSA}(\mathrm{LN}(S_0)), \\
S_0' &= \widetilde S_0 + \mathrm{FFN}(\mathrm{AdaLN}(\widetilde S_0, z^{\mathrm{ego}})).
\end{aligned}
\label{eq1}
\end{equation}
The future slots are predicted by a residual update $\hat S_{T_f} = S_0 + W_o(\mathrm{LN}(S_0'))$, which imposes a smooth transition prior while allowing the future object state to be conditioned on ego behavior. For optimization, latent action quantization is regularized by the VQ commitment loss $\mathcal{L}_{\mathrm{act}}$, while future slot prediction and target JEPA token reconstruction are supervised with mean-squared error losses, denoted as $\mathcal{L}_{\mathrm{object}}$ and $\mathcal{L}_{\mathrm{visual}}$, respectively. The overall Stage II loss is
\begin{equation}
    \mathcal{L}_{\mathrm{II}}
    =
    \lambda_{\mathrm{act}}\mathcal{L}_{\mathrm{act}}
    +
    \lambda_{\mathrm{object}}\mathcal{L}_{\mathrm{object}}
    +
    \lambda_{\mathrm{visual}}\mathcal{L}_{\mathrm{visual}}.
    \label{eq:stage2-loss}
\end{equation}
Together, Stages I and II constitute a self-supervised ego-object world model that infers latent ego behavior from history and predicts action-conditioned scene evolution.

\subsection{Interaction-Guided Trajectory Planning}

Stage III transfers the learned object-centric world model to supervised trajectory planning. To isolate the contribution of the learned world model, we use a lightweight DETR-style trajectory decoder as the planning head.
Given the observed history, learnable trajectory, action, and future-slot queries interact through self-attention and attend to scene context $C_t$ through cross-attention, where the context consists of dense visual tokens, current slots from the frozen recurrent slot encoder, and ego-condition tokens encoding ego state and route command.
The ego condition $e_t$ further modulates the decoder through adaptive layer normalization:
\begin{equation}
Q^{\ell+1}
=
\mathrm{Dec}^{\ell}
\bigl(
Q^{\ell}, C_t; e_t
\bigr),
\quad
\ell=1,\dots,L_{\mathrm{dec}}.
\label{eq2}
\end{equation}
The decoded queries produce $M$ trajectory hypotheses together with their corresponding latent actions, future slots, and confidence scores, which allows trajectory prediction to be guided by the inferred ego action while remaining aware of the corresponding object-level future scene evolution.

We train the planner with trajectory imitation and world-model feature alignment.
For multimodal trajectory prediction, we adopt a Winner-Takes-All objective, where only the trajectory hypothesis closest to the ground-truth ego trajectory is supervised with an $\ell_1$ imitation loss $\mathcal{L}_{\mathrm{traj}}$, while the predicted latent action and future slots of the same mode are aligned in feature space with frozen world model targets using mean-squared error losses $\mathcal{L}_{\mathrm{act}}$ and $\mathcal{L}_{\mathrm{slot}}$. Specifically, the latent action target is inferred from the complete training clip by the frozen Stage II tokenizer, whereas the slot target is obtained by encoding the ground-truth future frame with the frozen Stage I slot encoder. These targets are used only during training. Following prior candidate-selection practices~\cite{drivor}, candidate confidence is optionally supervised with dataset-specific oracle scores when available using $\mathcal{L}_{\mathrm{score}}$, and the highest-confidence trajectory is selected at inference time.
The overall Stage III objective is
\begin{small}
\begin{equation}
\mathcal{L}_{\mathrm{III}}
=
\lambda_{\mathrm{traj}}
\mathcal{L}_{\mathrm{traj}}
+
\lambda_{\mathrm{act}}
\mathcal{L}_{\mathrm{act}}
+
\lambda_{\mathrm{slot}}
\mathcal{L}_{\mathrm{slot}}
+
\lambda_{\mathrm{score}}
\mathcal{L}_{\mathrm{score}}.
\label{eq3}
\end{equation}
\end{small}
This objective enables the planner to adapt to dataset-specific trajectory styles while preserving the latent action and object-dynamics structure learned from raw videos.

Further architectural and implementation details of all three stages are provided in Appendix A.

%% file: exp.tex
\section{Experiments}


\paragraph{Experimental setup.}
\OURS uses only front-view images throughout all three stages. Frozen DINOv2 ViT-B/14 and V-JEPA2 ViT-g/16 serve as the image and video encoders, respectively, while a pretrained SDXL VAE~\cite{sdxl} provides appearance-level latent supervision and supports image reconstruction from predicted slots. We use $K=7$ 128-dimensional slots and a three-level residual action codebook containing 32, 64, and 128 codes of dimension 32. The planner predicts 64 trajectory hypotheses with confidence scores following~\cite{drivor}. All latent features have a unified dimension of 512, and the feed-forward layers use an expansion ratio of 4. Stages I-II are trained for 10 epochs and Stage III for 25 epochs on 8 NVIDIA A100 GPUs, using AdamW with a learning rate of $1\times10^{-4}$ and cosine annealing. The final Stage III planner contains approximately 60M trainable parameters. Unless otherwise specified, these settings are shared across all experiments. For NAVSIM, Stages I-II use \texttt{navtrain} and 40k
PhysicalAI-AV clips for training, whereas Stage III uses only
\texttt{navtrain}; for nuScenes, all stages use the official
training split.

\subsection{Benchmarks}

\paragraph{NAVSIM-v1.}

\begin{table}[t]
\centering
\fontsize{7.5pt}{8.5pt}\selectfont
\setlength{\tabcolsep}{1.8pt}
\renewcommand{\arraystretch}{1.0}
\caption{
\textbf{NAVSIM-v1} benchmark. We compare \OURS with published and recent methods on the test set (\texttt{navtest}). Higher is better for all metrics.
}
\label{tab:navsim_v1}
\begin{tabular*}{\columnwidth}{@{\hspace{2pt}\extracolsep{\fill}}l|ccccc|>{\columncolor{gray!18}}c@{\hspace{2pt}}}
\toprule
\textbf{Method} & NC$\uparrow$ & DAC$\uparrow$ & TTC$\uparrow$ & C$\uparrow$ & EP$\uparrow$ & \textbf{PDMS}$\uparrow$ \\
\midrule
PDM-Closed~\cite{pdmclosed}
& 94.6 & 99.8 & 89.9 & 86.9 & 99.9 & 89.1 \\
Human driver~\cite{navsim}
& 100 & 100 & 100 & 99.9 & 87.5 & 94.8 \\
\midrule
UniAD~\cite{uniad}
& 97.8 & 91.9 & 92.9 & \textbf{100} & 78.8 & 83.4 \\
PARA-Drive~\cite{paradrive}
& 97.9 & 92.4 & 93.0 & 99.8 & 79.3 & 84.0 \\
VAD-v2~\cite{vadv2}
& 98.1 & 94.8 & 94.3 & \textbf{100} & 80.6 & 86.2 \\
AutoVLA~\cite{autovla}
& 98.4 & 95.6 & \textbf{98.0} & 99.9 & 81.9 & 89.1 \\
GoalFlow~\cite{goalflow}
& 98.4 & 98.3 & 94.6 & \textbf{100} & 85.0 & 90.3 \\
ReCogDrive~\cite{recogdrive}
& 97.9 & 97.3 & 94.9 & \textbf{100} & 87.3 & 90.8 \\
Hydra-MDP++~\cite{hydra}
& 98.6 & 98.6 & 95.1 & \textbf{100} & 85.7 & 91.0 \\
DiffusionDriveV2~\cite{diffusiondrivev2}
& 98.3 & 97.9 & 94.8 & 99.9 & 87.5 & 91.2 \\
iPad~\cite{ipad}
& 98.6 & 98.3 & 94.9 & \textbf{100} & 88.0 & 91.7 \\
SparseDriveV2~\cite{sparsedrivev2}
& 98.5 & 98.4 & 95.0 & 99.9 & 88.6 & 92.0 \\
Drive-JEPA~\cite{drivejepa}
& 99.1 & 98.2 & 95.9 & 99.9 & 90.8 & 93.3 \\
DriveSuprim~\cite{drivesuprim}
& 98.6 & 98.6 & 95.5 & \textbf{100} & \textbf{91.3} & 93.5 \\
DrivoR~\cite{drivor}
& 99.0 & 98.9 & 96.7 & \textbf{100} & 90.0 & 93.7 \\
\midrule
LAW~\cite{law}
& 96.4 & 95.4 & 88.7 & 99.9 & 81.7 & 84.6 \\
World4Drive~\cite{world4drive}
& 97.4 & 94.3 & 92.8 & \textbf{100} & 79.9 & 85.1 \\
DriveLaW~\cite{drivelaw}
& 99.0 & 97.1 & 96.7 & \textbf{100} & 81.3 & 89.1 \\
DriveVLA-W0~\cite{drivevlaw0}
& 98.7 & \textbf{99.1} & 95.3 & 99.3 & 83.3 & 90.2 \\
EponaV2~\cite{eponav2}
& 98.6 & 97.9 & 95.7 & \textbf{100} & 84.8 & 90.4 \\
GeoWorldAD~\cite{geoworldad}
& 99.0 & 97.8 & 95.8 & 99.9 & 85.9 & 91.0 \\
\midrule
\textbf{INTERACT (ours)}
& \textbf{99.3} & \textbf{99.1} & 97.3 & \textbf{100} & 89.3 & \textbf{93.8} \\
\bottomrule
\end{tabular*}
\end{table}

NAVSIM~\cite{navsim} is a real-world planning benchmark built upon OpenScene~\cite{openscene} and nuPlan~\cite{nuplan}. NAVSIM evaluates predicted trajectories using a simulation-based protocol and reports closed-loop-inspired rule-based metrics, including no at-fault collision (NC), drivable area compliance (DAC), time-to-collision with bounds (TTC), ego progress (EP), and comfort (C). These metrics are further aggregated into the primary metric, Predictive Driver Model Score (PDMS). Notably, as shown in Table~\ref{tab:navsim_v1}, \OURS outperforms recent planning methods, including
world-model-based approaches such as DriveLaW~\cite{drivelaw} and EponaV2~\cite{eponav2}, particularly on safety- and compliance-related metrics such as NC, DAC, and TTC. This highlights the advantage of structured object-centric dynamics in providing informative scene-level representations for trajectory planning.

\paragraph{NAVSIM-v2.}

\begin{table*}[t]
\centering
\footnotesize
\setlength{\tabcolsep}{3.8pt}
\renewcommand{\arraystretch}{1.0}
\caption{
Results on the \textbf{NAVSIM-v2 \texttt{navtest}} benchmark. EPDMS$^{*}$ denotes results computed with the original NAVSIM-v2 evaluation code before applying the human-behavior filtering fix, while EPDMS denotes results computed with the updated official implementation. Higher values indicate better performance for all metrics.
}
\label{tab:navsim_v2}
\begin{tabular*}{\textwidth}{@{\hspace{4pt}\extracolsep{\fill}}l|ccccccccc|>{\columncolor{gray!18}}c>{\columncolor{gray!28}}c@{\hspace{4pt}}}
\toprule
\textbf{Method}
& NC$\uparrow$
& DAC$\uparrow$
& DDC$\uparrow$
& TL$\uparrow$
& EP$\uparrow$
& TTC$\uparrow$
& LK$\uparrow$
& HC$\uparrow$
& EC$\uparrow$
& \textbf{EPDMS}$^{*}\uparrow$
& \textbf{EPDMS}$\uparrow$ \\
\midrule
ReCogDrive~\cite{recogdrive}
& 98.3 & 95.2 & 99.5 & 99.8 & 87.1 & 97.5 & 96.6 & 98.3 & 86.5 & 83.6 & -- \\
Hydra-MDP++~\cite{hydra}
& 98.4 & 98.0 & 99.4 & 99.8 & 87.5 & 97.7 & 95.3 & 98.3 & 77.4 & 85.1 & -- \\
DiffusionDriveV2~\cite{diffusiondrivev2}
& 97.7 & 96.6 & 99.2 & 99.8 & 88.9 & 97.2 & 96.0 & 97.8 & \textbf{91.0} & 85.5 & 87.5 \\
SparseDriveV2~\cite{sparsedrivev2}
& 98.1 & 98.1 & 99.6 & 99.8 & \textbf{91.1} & 97.3 & 96.9 & 98.2 & 78.4 & 86.7 & 90.1 \\
DriveSuprim~\cite{drivesuprim}
& 98.4 & 98.6 & 99.6 & 99.8 & 90.5 & 97.8 & 97.0 & 98.3 & 78.6 & 87.1 & -- \\
\midrule
World4Drive~\cite{world4drive}
& 97.8 & 96.3 & 99.4 & 99.8 & 88.3 & 97.1 & \textbf{97.7} & 98.0 & 53.9 & 84.8 & -- \\
DriveVLA-W0~\cite{drivevlaw0}
& 98.5 & \textbf{99.1} & 98.0 & 99.7 & 86.4 & 98.1 & 93.2 & 97.9 & 58.9 & -- & 86.1 \\
DriveLaW~\cite{drivelaw}
& 98.7 & 96.9 & 99.6 & 99.8 & 87.5 & 98.3 & 97.6 & \textbf{98.4} & 77.4 & -- & 88.6 \\
EponaV2~\cite{eponav2}
& 98.5 & 97.4 & 99.5 & \textbf{99.9} & 87.9 & 98.1 & \textbf{97.7} & 98.2 & 77.4 & -- & 88.9 \\
GeoWorldAD~\cite{geoworldad}
& 99.0 & 97.8 & 99.6 & 99.7 & 89.1 & 98.6 & 97.6 & 98.0 & 82.2 & -- & 90.4 \\
\midrule
\textbf{INTERACT (ours)}
& \textbf{99.3} & 99.0 & \textbf{99.7} & 99.7 & 90.9 & \textbf{99.0} & 96.3 & 98.3 & 74.5 & \textbf{87.8} & \textbf{91.4} \\
\bottomrule
\end{tabular*}
\end{table*}

NAVSIM-v2~\cite{pseudo} extends NAVSIM-v1 with richer rule-compliance and driving-quality metrics, including driving
direction compliance (DDC), traffic light compliance (TL), lane
keeping (LK), history comfort (HC), and extended comfort (EC), as well as scene perturbations generated with Gaussian Splatting. We evaluate on both \texttt{navtest} and \texttt{navhard}. The former evaluates broad generalization across diverse scenes, whereas \texttt{navhard} uses real safety-critical scenes and synthetic counterparts for two-stage closed-loop evaluation under distribution shift. As shown in Table~\ref{tab:navsim_v2}, \OURS achieves the highest official Extended PDMS (EPDMS) and the best EPDMS$^{*}$ on \texttt{navtest}. Due to space limitations, the results on \texttt{navhard} are provided in Appendix C. Overall, our method demonstrates strong performance under the
extended safety, compliance, and robustness evaluation protocols.


\paragraph{NuScenes.}

\begin{table}[t]
\centering
\setlength{\tabcolsep}{3.8pt}
\renewcommand{\arraystretch}{1.0}
\caption{
Open-loop planning results on \textbf{nuScenes}.
}
\label{tab:nuscenes_planning}

\resizebox{\linewidth}{!}{
\begin{tabular}{lcc}
\toprule
\textbf{Method}
& \textbf{L2 Avg. (m)}$\downarrow$
& \textbf{Collision Avg. (\%)}$\downarrow$ \\
\midrule

UniAD~\cite{uniad}
& 0.69 & 0.12 \\


PARA-Drive~\cite{paradrive}
& 0.48 & 0.25 \\


AutoVLA~\cite{autovla}
& 0.40 & 0.20 \\

OpenREAD~\cite{openread}
& 0.36 & 0.11 \\

ORION~\cite{orion}
& 0.34 & 0.37 \\

OpenDrive-VLA~\cite{opendrivevla}
& 0.33 & 0.10 \\

OmniDrive~\cite{omnidrive}
& 0.33 & 0.30 \\

DriveTransformer-Large~\cite{drivetransformer}
& 0.33 & \textbf{0.07} \\

EMMA~\cite{emma}
& \textbf{0.32} & -- \\

AutoMoT~\cite{automot}
& \textbf{0.32} & \textbf{0.07} \\

\midrule

\textbf{INTERACT (ours)}
& \textbf{0.32} & 0.10 \\

\bottomrule
\end{tabular}
}
\end{table}

As a complementary open-loop evaluation, we further conduct experiments on the nuScenes benchmark~\cite{nuscenes}. We adopt a single-mode trajectory planning setting and report the prediction accuracy and collision rate of the planned ego trajectory. Due to space constraints, Table~\ref{tab:nuscenes_planning} reports the average results across the evaluated horizons, while the complete per-horizon results are provided in Appendix C. Our method achieves strong planning performance, demonstrating its effectiveness beyond the NAVSIM evaluation protocol. Notably, \OURS remains competitive with recent driving models. More details about datasets and metrics are in Appendix B.

\subsection{Visualization and Analysis}

\paragraph{Object-Centric World Model Diagnostics.}
\begin{figure*}[t]
\centering
\includegraphics[width=0.96\textwidth]{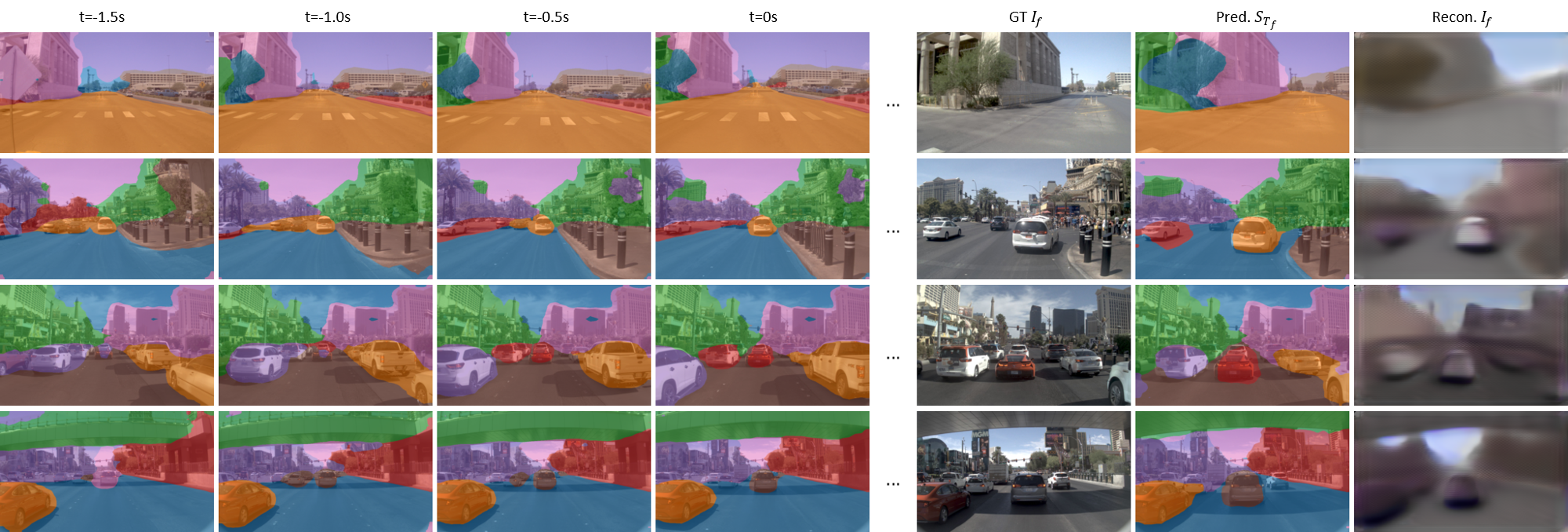}
\caption{
Visualization of object-centric slot dynamics and future prediction. The first four columns show historical slot assignments extracted by the Stage I recurrent slot encoder. The last three columns show the ground-truth future image, the future slot layout predicted by the Stage II conditional slot predictor, and the reconstructed future image, respectively.
}
\label{fig_slot_vis}
\end{figure*}

Before analyzing the contribution of the learned world model to planning, we first visually inspect whether it captures meaningful object-centric scene dynamics. Figure~\ref{fig_slot_vis} shows historical slot assignments produced by the Stage I recurrent slot encoder, the future slot layout  predicted by the
Stage II conditional slot predictor, and the corresponding image reconstruction. The colored maps represent pixel-to-slot assignments, with each color indicating the visual regions associated with a latent slot. Across consecutive frames, slot assignments exhibit temporal coherence over scene entities and broader regions. Conditioned on the current slots and inferred ego action, the model further predicts future layouts that capture changes in surrounding participants and scene geometry. Although the slot-based reconstructions omit fine-grained appearance details, reflecting the highly compressed nature of the slot representation, they preserve the dynamic structure most relevant to scene understanding and planning. Additional quantitative slot diagnostics, an evaluation of the ego-motion information recoverable from latent actions, and qualitative visualizations are provided in Appendix C.

\paragraph{Case Analysis.}
Figure~\ref{fig:hard_case_vis} presents three NAVSIM scenes. The upper and lower rows show the highest- and lowest-confidence modes ranked by the scorer, respectively. Predicted and ground-truth trajectories are red and green.
In Scenario A and B, the highest-ranked mode adopts a conservative trajectory that maintains a larger distance from the leading vehicle, whereas the aggressive alternative closes the gap excessively. Its future slot prediction indicates a reduced ego-relative separation and thus collision risk. In Scenario C, the selected trajectory follows the road geometry and closely matches the ground truth, while its future slots remain consistent with the scene evolution. In contrast, the lowest-ranked alternative deviates substantially from the drivable region, with the corresponding slot prediction reflecting the off-road deviation. These examples suggest that mode-conditioned slot prediction captures both interaction risk and spatial feasibility, providing informative cues for trajectory selection.

\begin{figure}[t]
\centering
\includegraphics[width=\linewidth]{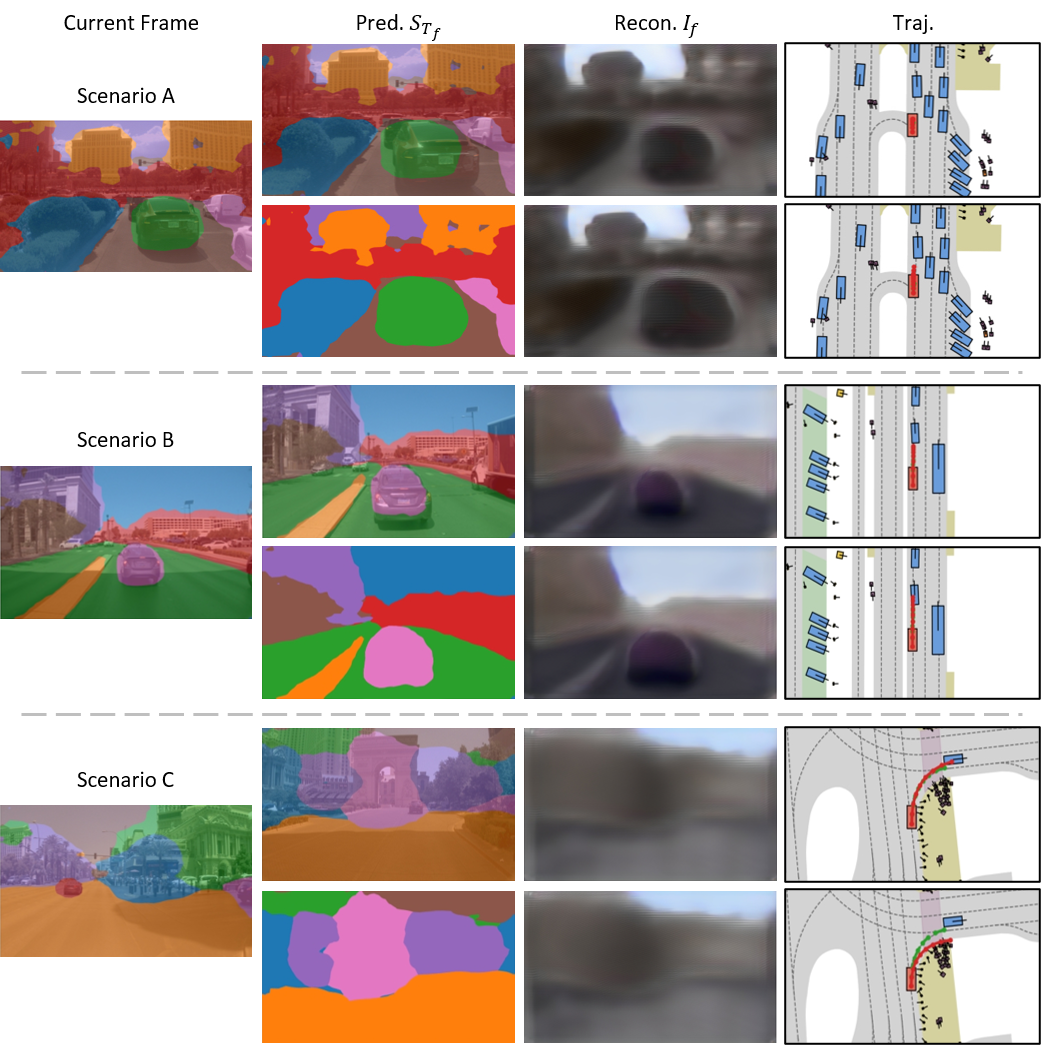}
\caption{
Mode-wise planning and future-slot predictions in three NAVSIM scenes. The upper and lower rows show the highest- and lowest-ranked modes, respectively. Red and green denote predicted and ground-truth trajectories.
}
\label{fig:hard_case_vis}
\end{figure}

\subsection{Ablation Study}

Unless otherwise specified, all ablation experiments are trained with the NAVSIM \texttt{navtrain} split and evaluated on the \texttt{navtest} set using a 16-mode planner for efficient ablation evaluation.

\paragraph{What Contributes to Planning?}
We ablate the three world-model signals transferred to Stage III: current-slot context, latent-action alignment, and future-slot alignment. As shown in Table~\ref{tab:ablation_components}, each component brings consistent overall gains over the trajectory-only baseline. Introducing slot context improves PDMS from 90.9 to
91.3, showing the benefit of an object-centric scene representation. Adding latent-action alignment further raises PDMS to 91.6, while future-slot alignment yields a larger improvement, indicating that anticipated object dynamics provide useful guidance for planning. Combining all three signals achieves the best PDMS and the highest TTC, confirming their complementary contributions
to safe and accurate trajectory prediction.






\begin{table}[t]
\centering
\caption{
Ablation of the world-model signals transferred to Stage III on NAVSIM \texttt{navtest}. A common frozen Stage I-II teacher is used wherever alignment targets are required.
}
\label{tab:ablation_components}
\resizebox{\linewidth}{!}{
\begin{tabular}{ccccccc}
\toprule
Slot Context
& Action Align.
& Future-Slot Align.
& NC$\uparrow$ & DAC$\uparrow$ & TTC$\uparrow$ & PDMS$\uparrow$ \\
\midrule
\xmark & \xmark & \xmark
& 98.9 & 97.6 & 96.4 & 90.9 \\

\cmark & \xmark & \xmark
& 99.2 & 97.7 & 97.0 & 91.3 \\

\cmark & \cmark & \xmark
& 99.2 & 98.0 & 97.1 & 91.6 \\

\cmark & \xmark & \cmark
& \textbf{99.3} & \textbf{98.4} & 97.1 & 92.0 \\

\cmark & \cmark & \cmark
& \textbf{99.3} & \textbf{98.4} & \textbf{97.3} & \textbf{92.2} \\
\bottomrule
\end{tabular}
}
\end{table}

\paragraph{Does Action Conditioning Matter?}
We compare the full action-conditioned slot predictor with a variant that predicts future slots independently of the latent ego action. As shown in Table~\ref{tab:ablation_action_conditioning}, action conditioning improves PDMS from 91.9 to 92.2 with only a marginal decrease in EP. This result validates that coupling ego actions with future slot evolution promotes safer and more compliant planning, while keeping the learned action representation consistent with the resulting environment dynamics, leading to more coherent ego-environment modeling and better downstream planning.

\begin{table}[t]
\centering
\renewcommand{\arraystretch}{0.9}
\caption{
Effect of ego-action conditioning in Stage II.
}
\label{tab:ablation_action_conditioning}
\resizebox{\linewidth}{!}{
\begin{tabular}{cccccc}
\toprule
Action Conditioning
& NC$\uparrow$
& DAC$\uparrow$
& TTC$\uparrow$
& EP$\uparrow$
& PDMS$\uparrow$ \\
\midrule

\xmark
& 99.2 & 98.3 & 97.0 & \textbf{86.4} & 91.9 \\

\cmark
& \textbf{99.3} & \textbf{98.4} & \textbf{97.3} & 86.2 & \textbf{92.2} \\
\bottomrule
\end{tabular}
}
\end{table}

\paragraph{Watch more, drive better?}
We study how the scale of self-supervised video pretraining affects downstream planning. The trajectory-only configuration is identical to the first row of Table~\ref{tab:ablation_components}. All pretrained variants use the same Stage III model and training configuration, while only the video data used to pretrain Stages I and II are varied.

\begin{table}[t]
\centering
\renewcommand{\arraystretch}{0.9}
\caption{
Scaling self-supervised video pretraining.
}
\label{tab:data_scaling}
\resizebox{\linewidth}{!}{
\begin{tabular}{lccccc}
\toprule
Stage I-II Video Data
& NC$\uparrow$
& DAC$\uparrow$
& TTC$\uparrow$
& EP$\uparrow$
& PDMS$\uparrow$ \\
\midrule
Trajectory-only (no Stage I-II)
& 98.9 & 97.6 & 96.4 & 85.6 & 90.9 \\

20k \texttt{navtrain}
& 99.2 & 98.0 & 97.1 & 86.2 & 91.7 \\

50k \texttt{navtrain}
& 99.2 & 98.2 & 97.2 & 86.1 & 91.9 \\

\texttt{navtrain}
& 99.3 & 98.4 & \textbf{97.3} & 86.2 & 92.2 \\

\texttt{navtrain} + 40k PhysicalAI-AV
& \textbf{99.4} & \textbf{98.5} & \textbf{97.3} & 86.3 & \textbf{92.4} \\


\bottomrule
\end{tabular}
}
\end{table}


Table~\ref{tab:data_scaling} shows that increasing the amount of in-domain pretraining data consistently improves downstream planning performance, indicating that broader video coverage enables the world
model to learn more transferable ego-object dynamics. Incorporating 40k external PhysicalAI-AV clips~\cite{cosmos} further raises PDMS, suggesting that additional unlabeled driving videos provide complementary dynamics beyond the target dataset. Nevertheless, with limited pretraining data, the teacher may not capture sufficiently diverse driving behaviors and scene transitions, which can make its supervision less effective than some simpler transfer configurations in
Table~\ref{tab:ablation_components}. Overall, all pretrained variants outperform the trajectory-only baseline, confirming the effectiveness of self-supervised world-model pretraining and the positive impact of scaling pretraining data.

%% file: conclu.tex
\section{Conclusion}
We present \OURS, a framework that learns interaction-aware planning representations directly from raw driving videos. Through self-supervised object-centric and action-conditioned world modeling, \OURS captures structured ego--environment dynamics without object-level annotations and transfers them to an end-to-end trajectory planner. The resulting scene structure and ego-conditioned future cues support trajectory generation and selection. Experiments across standard driving benchmarks, together with ablations and visualizations, demonstrate strong planning performance and validate the learned interaction dynamics.

%% file: appen.tex
\section{Appendix}

\subsection{A. Additional Architecture Details}
\label{app}
This section supplements the structural details omitted from the main text.

\paragraph{Stage I: Object-centric decoders.}
Input images are resized to $518\times518$ and encoded into a $37\times37$ patch grid by the frozen DINOv2 backbone. SlotMixer uses learned patch queries and three Transformer allocation blocks to recover a patch-aligned representation from the unordered slot set. Separate prediction heads reconstruct the DINO features and estimate temporal correspondences between adjacent frames. The patch-to-slot allocation learned within SlotMixer is independent of the soft assignment masks produced by recurrent Slot Attention.

For appearance supervision, the Slot Attention masks spatially compose the slot features into a dense feature map. This representation is resized to the SDXL latent resolution and processed by three convolutional blocks followed by an output projection, producing a latent compatible with the frozen SDXL VAE.

\paragraph{Stage II: Motion tokenizer and slot prediction.}
V-JEPA2 processes video clips at a resolution of $224\times224$. In the NAVSIM setting, each clip contains four historical and eight future frames sampled at 2Hz, and the tokens from the final temporal group are used as the future prediction target.

MFormer introduces a learned motion query that jointly attends to the context and target tokens through spatiotemporal deformable cross-attention. The resulting sequence is processed by four Transformer encoder layers, and the output motion token is further refined through three cross-attention layers over the historical slots before residual vector quantization.

The conditional slot predictor performs one-shot prediction at the final future horizon rather than autoregressive prediction over all future frames. It consists of four predictor blocks, each combining slot self-attention with an action-conditioned feed-forward layer using adaptive normalization. The final future slots are predicted as residual updates to the current slot state. A four-layer Transformer decoder is additionally used for the V-JEPA2 reconstruction objective.

\paragraph{Stage III: Mode-wise trajectory planner.}
Each of the 64 trajectory modes contains one trajectory query, three action queries corresponding to the residual codebook levels, and seven future-slot queries. The future-slot queries are initialized from the current slots, whereas the trajectory and action queries are learned. Queries from all modes are processed jointly by a shared six-layer DETR decoder.

\paragraph{Training hyperparameters.}
All three training stages use the AdamW optimizer with a batch size of 16, an initial learning rate of 1e-4, and a weight decay of 0.01. We adopt a learning-rate warm-up over the first 10\% of the training schedule, followed by cosine decay. Unless otherwise specified, these hyperparameters are shared across all three stages.Table~\ref{tab:loss_weights} summarizes the loss coefficients used
for the three training stages. Unless otherwise specified, the same
weights are used across all experiments. For datasets where
candidate-level oracle scores are unavailable, the scoring loss is
disabled by setting $\lambda_{\mathrm{score}}=0$.

\begin{table}[t]
\centering
\footnotesize
\setlength{\tabcolsep}{5.0pt}
\renewcommand{\arraystretch}{1.0}
\caption{
Loss weights used in the three training stages.
The scoring loss is enabled only when candidate-level oracle
supervision is available.
}
\label{tab:loss_weights}
\resizebox{\linewidth}{!}{
\begin{tabular}{l|lc}
\toprule
\textbf{Stage} & \textbf{Loss term} & \textbf{Weight} \\
\midrule

\multirow{3}{*}{Stage I}
& DINO feature reconstruction $\mathcal{L}_{\mathrm{feat}}$
& $\lambda_{\mathrm{feat}}=\mathrm{1.0}$ \\
& Temporal correspondence $\mathcal{L}_{\mathrm{temp}}$
& $\lambda_{\mathrm{temp}}=\mathrm{1.0}$ \\
& VAE latent reconstruction $\mathcal{L}_{\mathrm{vae}}$
& $\lambda_{\mathrm{vae}}=\mathrm{1.0}$ \\

\midrule

\multirow{3}{*}{Stage II}
& Latent-action quantization $\mathcal{L}_{\mathrm{act}}$
& $\lambda_{\mathrm{act}}^{\mathrm{II}}=\mathrm{1.0}$ \\
& Future-slot prediction $\mathcal{L}_{\mathrm{object}}$
& $\lambda_{\mathrm{object}}=\mathrm{1.0}$ \\
& JEPA feature reconstruction $\mathcal{L}_{\mathrm{visual}}$
& $\lambda_{\mathrm{visual}}=\mathrm{1.0}$ \\

\midrule

\multirow{4}{*}{Stage III}
& Trajectory imitation $\mathcal{L}_{\mathrm{traj}}$
& \begin{tabular}[c]{@{}c@{}}
$\lambda_{\mathrm{traj}}=1.0$ (NAVSIM)\\
$\lambda_{\mathrm{traj}}=5.0$ (nuScenes)
\end{tabular} \\
& Latent-action alignment $\mathcal{L}_{\mathrm{act}}$
& $\lambda_{\mathrm{act}}^{\mathrm{III}}=1.0$ \\
& Future-slot alignment $\mathcal{L}_{\mathrm{slot}}$
& $\lambda_{\mathrm{slot}}=1.0$ \\
& Candidate scoring $\mathcal{L}_{\mathrm{score}}$
& $\lambda_{\mathrm{score}}=1.0$ \\

\bottomrule
\end{tabular}
}
\end{table}

\begin{figure*}[t]
\centering
\includegraphics[width=0.96\textwidth]{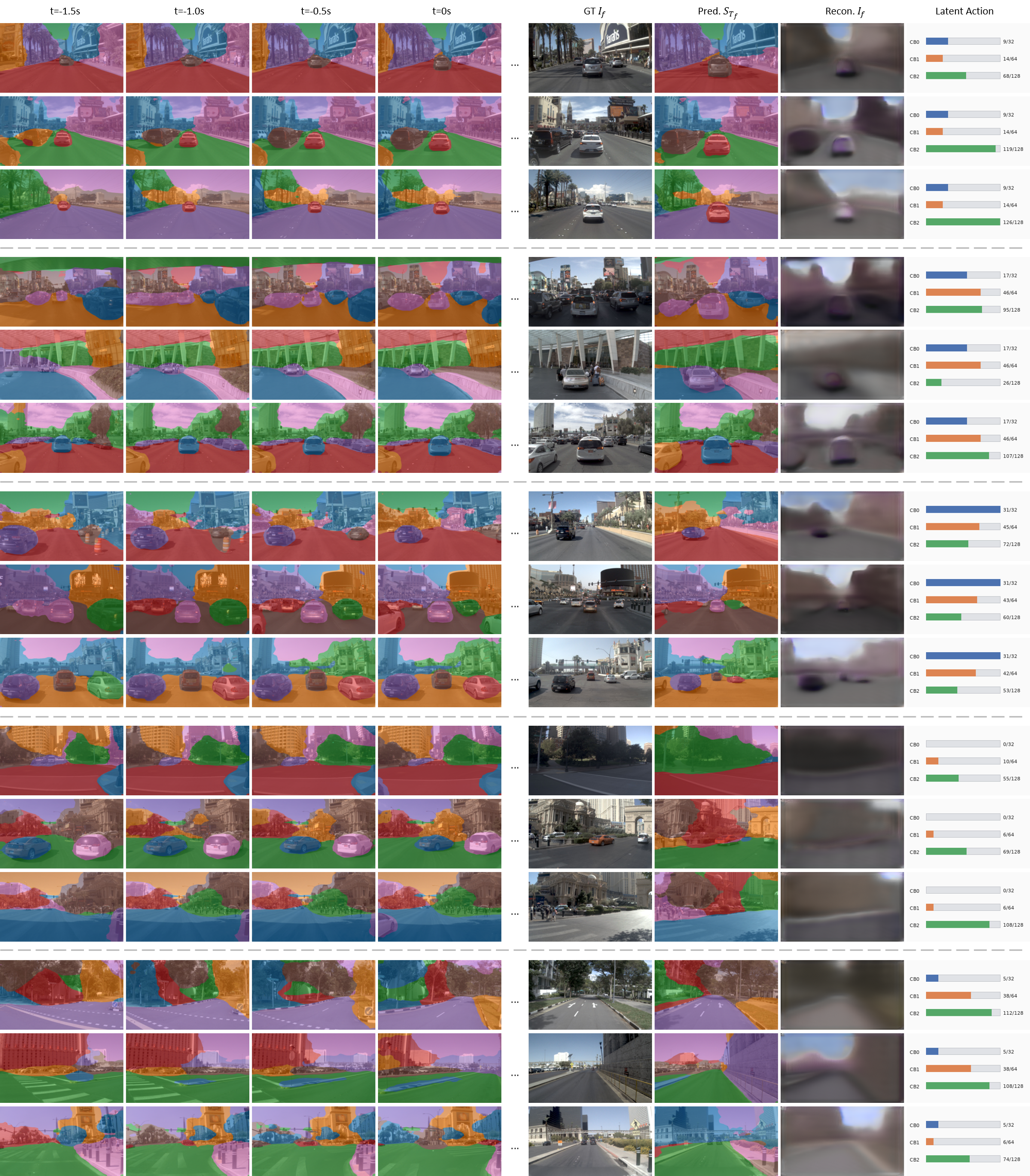}
\caption{
Additional visualizations of object-centric dynamics and learned latent actions. Each row shows historical slot assignments, the ground-truth future image, the predicted future slots, the slot-based future reconstruction, and the three-level residual codebook indices of the inferred ego action. Dashed lines group representative driving behaviors, including car following, approaching and stopping behind a leading vehicle, accelerating from an intersection, and right-turn maneuvers. Samples within each group exhibit similar latent-action codes, with identical assignments at the first quantization level, indicating that meaningful coarse ego-motion patterns emerge without action supervision.
}
\label{fig_slotaction}
\end{figure*}

\subsection{B. More Details on Datasets and Metrics}
\paragraph{NAVSIM.}
NAVSIM~\cite{navsim} is built upon the OpenScene redistribution of nuPlan and retains challenging planning scenarios sampled at 2\,Hz. The standard \texttt{navtrain} and \texttt{navtest} splits contain approximately 103K and 12K samples, respectively. NAVSIM-v1 evaluates predictions on \texttt{navtest}. 
Each predicted ego trajectory is rolled out over a four-second horizon in a non-reactive simulator, where surrounding agents follow their logged motions. The final score combines mandatory rule-compliance metrics with weighted driving-quality metrics:
\begin{equation}
\mathrm{Score}
=
\left(\prod_{m\in\mathcal{M}} s_m\right)
\left(
\frac{\sum_{a\in\mathcal{A}}\omega_a s_a}
{\sum_{a\in\mathcal{A}}\omega_a}
\right),
\label{eq:navsim_score}
\end{equation}
where $\mathcal{M}$ and $\mathcal{A}$ denote the multiplicative and weighted metric sets, respectively, and $s$ and $\omega$ denote a subscore and its associated weight. For NAVSIM-v1, $\mathcal{M}$ contains no at-fault collision (NC) and drivable area compliance (DAC), while $\mathcal{A}$ contains ego progress (EP), time-to-collision (TTC), and comfort (C), yielding the PDMS. NAVSIM-v2 extends the multiplicative set with driving direction compliance (DDC) and traffic light compliance (TL), and augments the weighted set with lane keeping (LK) and extended comfort (EC), while replacing C with history comfort (HC). Their aggregation produces the EPDMS.

\paragraph{NuScenes.}
NuScenes is a real-world autonomous driving dataset with diverse urban scenes and fine-grained annotations, and has been widely used to evaluate trajectory planning performance~\cite{nuscenes}. It contains 1,000 driving scenes collected in Boston and Singapore, with 700 scenes for training, 150 for validation, and 150 for testing. We follow the open-loop planning protocol introduced by ST-P3~\cite{stp3}.

\subsection{C. Supplementary Experiment Results}

\paragraph{Navsim-v2 \texttt{navhard-two-stage} benchmark.}

\begin{table*}[t]
\centering
\scriptsize
\setlength{\tabcolsep}{2.0pt}
\renewcommand{\arraystretch}{1.0}
\caption{
Results on the \textbf{NAVSIM-v2 \texttt{navhard}} benchmark.
Stage 1 evaluates real safety-critical scenes, while Stage 2
evaluates their Gaussian-Splatting-generated synthetic counterparts
under perturbed ego states. EPDMS denotes the final aggregated
two-stage score. Higher values indicate better performance for all
metrics.
}
\label{tab:navhard}

\begin{tabular*}{\textwidth}{
@{\hspace{2pt}\extracolsep{\fill}}
l|
ccccccccc|
ccccccccc|
c
@{\hspace{2pt}}
}
\toprule

\multirow{2}{*}{\textbf{Method}}
& \multicolumn{9}{c|}{\textbf{Stage 1}}
& \multicolumn{9}{c|}{\textbf{Stage 2}}
& \multirow{2}{*}{\textbf{EPDMS}$\uparrow$} \\

\cmidrule(lr){2-10}
\cmidrule(lr){11-19}

& NC$\uparrow$
& DAC$\uparrow$
& DDC$\uparrow$
& TL$\uparrow$
& EP$\uparrow$
& TTC$\uparrow$
& LK$\uparrow$
& HC$\uparrow$
& EC$\uparrow$

& NC$\uparrow$
& DAC$\uparrow$
& DDC$\uparrow$
& TL$\uparrow$
& EP$\uparrow$
& TTC$\uparrow$
& LK$\uparrow$
& HC$\uparrow$
& EC$\uparrow$

& \\

\midrule

ReCogDrive~\cite{recogdrive}
& 96.4 & 78.9 & 98.7 & 99.8 & 82.6 & 95.6 & 94.4 & 97.6 & 74.2
& 80.2 & 65.0 & 82.4 & 98.7 & 85.2 & 76.9 & 43.8 & 96.6 & 71.8
& 25.7 \\

DiffusionDrive~\cite{diffusiondrive}
& 96.8 & 86.0 & 98.8 & 99.3 & 84.0 & 95.8 & 96.7 & 97.6 & \textbf{79.6} 
& 80.1 & 72.8 & 84.4 & 98.4 & 85.9 & 76.6 & 46.4 & 96.3 & \textbf{72.8}
& 27.5 \\

DriveSuprim~\cite{drivesuprim}
& \textbf{98.9} & 95.1 & 99.2 & 99.6 & 76.1 & \textbf{99.1} & 94.7 & 97.6 & 54.2
& 87.9 & 88.8 & 89.6 & 98.8 & 80.3 & 86.0 & 53.5 & 97.1 & 56.1
& 42.1 \\

GuideFlow~\cite{guideflow}
& 97.8 & 97.1 & \textbf{100} & \textbf{100} & 81.4 & 98.5 & 91.4 & 92.8 & 34.2
& 87.3 & \textbf{92.3} & \textbf{98.0} & 96.9 & 75.8 & 85.5 & \textbf{59.3} & 95.4 & 53.5
& 46.7 \\

\midrule

DriveVLA-W0~\cite{drivevlaw0}
& 96.8 & 83.3 & 99.0 & 99.6 & \textbf{84.6} & 95.3 & 96.4 & 97.6 & 78.2
& 76.8 & 64.3 & 79.9 & 98.3 & \textbf{89.2} & 75.0 & 46.8 & 95.8 & 53.1
& 24.4 \\

DriveLaW~\cite{drivelaw}
& 97.3 & 89.1 & 99.2 & 99.6 & 84.3 & 97.1 & 96.2 & 97.8 & 67.6
& 82.5 & 67.6 & 83.5 & 98.1 & 84.8 & 78.5 & 45.8 & 96.4 & 57.3
& 30.6 \\

World4Drive~\cite{world4drive}
& 97.3 & 89.1 & 97.6 & 99.7 & 60.5 & 96.8 & 87.7 & 93.1 & 60.0 
& 91.4 & 82.0 & 91.0 & 98.5 & 53.1 & \textbf{90.6} & 52.3 & 93.3 & 62.8
& 34.9 \\

EponaV2~\cite{eponav2}
& 97.3 & 90.7 & 99.4 & 100 & 83.3 & 97.3 & \textbf{97.3} & 97.6 & 60.9
& 83.6 & 78.0 & 88.0 & 98.9 & 86.0 & 80.3 & 50.1 & 96.1 & 52.0
& 36.1 \\

\midrule

\textbf{INTERACT (ours)}
& \textbf{98.9} & \textbf{97.6} & 99.6
& \textbf{100} & 75.2 & \textbf{99.1}
& 94.2 & \textbf{97.8} & 51.1
& \textbf{93.0} & 85.7 & 92.4
& \textbf{99.5} & 71.5 & 90.5
& 50.6 & \textbf{97.6} & 62.1
& \textbf{48.1} \\

\bottomrule
\end{tabular*}
\end{table*}

\begin{table*}[t]
\centering
\footnotesize
\setlength{\tabcolsep}{3.8pt}
\renewcommand{\arraystretch}{1.0}
\caption{
Complete open-loop planning results on nuScenes.
Results are highlighted in \textbf{bold} and \underline{underlined} for the best and second-best performance among all compared methods, respectively.
Lower is better for all metrics.
}
\label{tab:nuscenes_full_results}
\begin{tabular*}{0.86\textwidth}{@{\hspace{4pt}\extracolsep{\fill}}l|cccc|cccc@{\hspace{4pt}}}
\toprule
\multirow{2}{*}{\textbf{Method}}
& \multicolumn{4}{c|}{\textbf{L2 (m)}$\downarrow$}
& \multicolumn{4}{c}{\textbf{Collision (\%)}$\downarrow$} \\
& 1s & 2s & 3s & Avg.
& 1s & 2s & 3s & Avg. \\
\midrule
UniAD~\cite{uniad}
& 0.44 & 0.67 & 0.96 & 0.69
& 0.04 & \underline{0.08} & 0.23 & 0.12 \\
ParaDrive~\cite{paradrive}
& 0.25 & 0.46 & 0.74 & 0.48
& 0.14 & 0.23 & 0.39 & 0.25 \\

AutoVLA~\cite{autovla}
& 0.21 & 0.38 & 0.60 & 0.40
& 0.13 & 0.18 & 0.28 & 0.20 \\
OpenREAD~\cite{openread}
& 0.17 & 0.34 & 0.56 & 0.36
& 0.04 & \underline{0.08} & 0.22 & 0.11 \\
ORION~\cite{orion}
& 0.17 & 0.31 & 0.55 & 0.34
& 0.05 & 0.25 & 0.80 & 0.37 \\
OpenDrive-VLA~\cite{opendrivevla}
& \underline{0.15} & 0.31 & 0.55 & \underline{0.33}
& \underline{0.01} & \underline{0.08} & 0.21 & \underline{0.10} \\
OmniDrive~\cite{omnidrive}
& \textbf{0.14} & \textbf{0.29} & 0.55 & \underline{0.33}
& \textbf{0.00} & 0.13 & 0.78 & 0.30 \\
DriveTransformer-Large~\cite{drivetransformer}
& 0.16 & \underline{0.30} & 0.55 & \underline{0.33}
& \underline{0.01} & \textbf{0.06} & \textbf{0.15} & \textbf{0.07} \\
EMMA~\cite{emma}
& \textbf{0.14} & \textbf{0.29} & \underline{0.54} & \textbf{0.32}
& -- & -- & -- & -- \\
AutoMoT~\cite{automot}
& \textbf{0.14} & \textbf{0.29} & \underline{0.54} & \textbf{0.32}
& \underline{0.01} & \textbf{0.06} & \textbf{0.15} & \textbf{0.07} \\
\midrule
\textbf{INTERACT (ours)}
& \underline{0.15} & \underline{0.30} & \textbf{0.52} & \textbf{0.32}
& 0.05 & 0.09 & \underline{0.16} & \underline{0.10} \\
\bottomrule
\end{tabular*}
\end{table*}

The complete results on the NAVSIM-v2 \texttt{navhard} benchmark are reported in Table~\ref{tab:navhard}. \texttt{Navhard} adopts a two-stage closed-loop evaluation protocol. Stage 1 evaluates planning performance on real safety-critical driving scenes, while Stage 2 evaluates the planner on corresponding
Gaussian-Splatting-generated synthetic scenes that reflect perturbed ego states and distribution shifts. As shown in Table~\ref{tab:navhard}, \OURS outperforms recent planning methods, with particularly clear gains over
world-model-based approaches.

\paragraph{Complete nuScenes planning results.}
Due to space constraints, the main paper reports only the average L2 displacement error and collision rate on nuScenes. Table~\ref{tab:nuscenes_full_results} provides the complete results at 1, 2, and 3 seconds. \OURS remains competitive  with other driving models and achieves particularly strong long-horizon trajectory accuracy.

\paragraph{Additional slot diagnostics.}
To examine whether the recurrent slot representation indeed groups scene entities into
spatially concentrated and temporally stable slots, we compare recurrent Slot Attention with
a generic learned-query baseline using two metrics: \emph{object-slot concentration} (OSC)
and \emph{track-slot consistency} (TSC). Both models use the same frozen DINOv2 ViT-B/14
features, $K=7$ slots, training objectives, data and training budget.
The baseline replaces Slot Attention with $K$ learnable queries decoded by a three-layer
Transformer decoder (self-attention, cross-attention to the patch tokens, FFN; 4 heads).
It therefore lacks the two key properties of Slot Attention:
(i) \emph{competition}, since its cross-attention is normalized over patches rather than
over slots, and
(ii) \emph{recurrence}, since every frame is decoded from the same queries rather than from
the previous frame's slots.

Let $A_t\in[0,1]^{P\times K}$ be the soft patch-to-slot assignment at frame $t$, with
$\sum_k A_t[p,k]=1$. For Slot Attention, $A_t$ is the last-iteration attention normalized
over slots. For the baseline, $A_t$ is the head-averaged last-layer cross-attention,
renormalized over queries for each patch. For an object $i$ whose projected 3D box covers
the patch set $\mathcal{P}_{i,t}$, we define
\begin{equation}
m_{i,t}^{k}=\frac{1}{|\mathcal{P}_{i,t}|}\sum_{p\in\mathcal{P}_{i,t}}A_t[p,k],
\qquad
k_{i,t}^{*}=\arg\max_{k} m_{i,t}^{k}.
\end{equation}
A track $i$ is an object followed within one sequence. Its valid frames $\mathcal{V}_i$
(object at least $60\%$ unoccluded and covering at least four patches, $T_i=|\mathcal{V}_i|\ge 3$)
yield
\begin{equation}
\mathrm{OSC}=\frac{1}{N}\sum_{i=1}^{N}\frac{1}{T_i}\sum_{t\in\mathcal{V}_i}\max_{k} m_{i,t}^{k},
\end{equation}
\begin{equation}
\mathrm{TSC}=\frac{1}{N}\sum_{i=1}^{N}\max_{k}\frac{1}{T_i}\sum_{t\in\mathcal{V}_i}
\mathbb{I}\left(k_{i,t}^{*}=k\right).
\end{equation}
OSC measures how much of an object's region falls into a single slot (lower bound $1/K=0.14$).
TSC measures how consistently the object keeps the same dominant slot over time (chance
level $0.44$ under our protocol).

We evaluate on the 12{,}146 NAVSIM navtest scenarios, each using the four history
front-camera frames observed by the planner. This yields $N=29{,}846$ vehicle and pedestrian
tracks. As shown in Table~\ref{tab:slot_diagnostics}, Slot Attention reaches 0.72 OSC and
0.86 TSC, whereas the learned queries stay near the respective lower bounds (0.22 and 0.58).
This indicates that slot competition and recurrence yield concentrated and temporally
consistent scene entities. These entities are region-level, e.g.\ nearby vehicles may share
a slot, rather than in strict one-to-one correspondence with objects.

\begin{table}[t]
\centering
\caption{Slot diagnostics on NAVSIM navtest (29{,}846 tracks).
OSC lower bound $1/K=0.14$; TSC chance level $0.44$.}
\label{tab:slot_diagnostics}
\begin{tabular}{lcc}
\hline
Representation & OSC $\uparrow$ & TSC $\uparrow$ \\
\hline
Generic Learned Queries  & 0.22          & 0.58          \\
Recurrent Slot Attention & \textbf{0.72} & \textbf{0.86} \\
\hline
\end{tabular}
\end{table}

\paragraph{Additional Latent-Action Analysis and World Model Visualizations.}
To investigate whether the learned latent actions retain physically meaningful information about ego motion, we conduct lightweight regression probes on three representations: pooled visual-transition features, pre-VQ latents, and quantized action embeddings. Under self-supervised training, the latent-action tokenizer is learned without explicit ground-truth trajectories or action labels. We use the regression probes as a post-hoc diagnostic to examine how much ego-motion information is decodable from the resulting representations.

Specifically, each regression probe is implemented as a lightweight
feed-forward network consisting of an input LayerNorm, FFN layers,
and a linear output projection. All probes use the same architecture and training configuration, with the input representations kept frozen. Given a representation
$\mathbf{h}_n^{(r)}$ from the $n$-th sample, the probe predicts
the corresponding ego-motion variables as
\begin{equation}
\hat{\mathbf{y}}_n^{(r)}
=
W_r^{\mathrm{out}}
\operatorname{FFN}_r
\left(
\operatorname{LN}(\mathbf{h}_n^{(r)})
\right)
+b_r^{\mathrm{out}},
\end{equation}

where $r$ denotes the input representation, $\mathbf{y}_n =
[\Delta x_n,\Delta\phi_n,\Delta v_n]$, including ego displacement, heading change, and speed change, respectively. The probes are evaluated on held-out samples, and their prediction errors are measured using the mean absolute error (MAE):
\begin{equation}
\mathrm{MAE}_{d}^{(r)}
=
\frac{1}{N_{\mathrm{test}}}
\sum_{n=1}^{N_{\mathrm{test}}}
\left|
\hat{y}_{n,d}^{(r)}-y_{n,d}
\right|,
\quad
d\in\{x,\phi,v\},
\end{equation}
where $N_{\mathrm{test}}$ denotes the number of held-out samples.

As reported in Table~\ref{tab:latent_action_probe}, the prediction errors consistently decrease from pooled visual-transition features to pre-VQ latents and further to quantized action tokens. In particular, quantized action tokens achieve MAEs of 0.9 m, 0.95 rad, and 2.0 m/s for ego displacement, heading change, and speed change, respectively. These results indicate that the learned discrete action representations retain decodable ego-motion information despite vector quantization, suggesting that the learned codes capture motion-related information beyond visual appearance alone.

\begin{table}[t]
\centering
\caption{Held-out regression probe results for ego-motion prediction from different representations. Lower MAE indicates better prediction accuracy.}
\label{tab:latent_action_probe}
\small
\resizebox{\linewidth}{!}{
\begin{tabular}{lccc}
\hline
Representation
& $\Delta x$ (m) $\downarrow$
& $\Delta\phi$ (rad) $\downarrow$
& $\Delta v$ (m/s) $\downarrow$ \\
\hline
Pooled visual-transition features
& 2.3 & 1.26 & 5.1 \\
Pre-VQ latents
& 1.4 & 1.04 & 2.6 \\
Quantized action tokens
& \textbf{0.9} & \textbf{0.95} & \textbf{2.0} \\
\hline
\end{tabular}}
\end{table}

We further provide qualitative results of the object-centric world model, together with the latent ego actions inferred by the Stage II tokenizer in  Figure~\ref{fig_slotaction}. Since the latent actions are learned without action annotations, individual codebook entries are not assigned predefined semantics and should not be interpreted as deterministic control commands. Nevertheless, representative samples reveal a clear behavior-dependent structure in the learned discrete action space.

From top to bottom, the visualized groups correspond approximately to car following, approaching and stopping behind a leading vehicle, accelerating from an intersection, and two sets of right-turn maneuvers. Scenes within the same behavior group produce highly similar residual code tuples. In particular, all examples within each group select the same code at the first, coarsest quantization level, while later levels capture finer variations caused by scene context, speed, and motion details. A behavior category is not necessarily represented by a unique code tuple, since multiple latent codes may describe similar maneuvers under different environments. However, the observed within-group consistency indicates that the self-supervised action tokenizer organizes visual transitions according to meaningful ego-motion patterns rather than merely encoding static appearance.

\paragraph{Additional Controlled Planning Ablations.}
We conduct additional controlled experiments to further investigate the contributions of future-slot supervision, scorer supervision, and world-model transfer to planning performance. Specifically, we compare structured future-slot alignment against dense feature alignment, evaluate world-model transfer under different scorer supervision strategies, and examine the effect of trajectory mode count. All experiments are trained on the NAVSIM navtrain split and evaluated using PDMS, with higher scores indicating better planning performance. For computational efficiency, all planning experiments use the 16-mode configuration unless otherwise specified.

Dense future-feature alignment.
\textit{Dense future-feature alignment.}
To examine whether structured future-slot supervision provides benefits beyond generic feature prediction, we compare dense future DINO-feature alignment with future-slot alignment. Both variants use slot context without action alignment, and differ only in their future alignment targets. As reported in Table~\ref{tab:dense_future_alignment}, dense future-feature alignment achieves 91.5 PDMS, while future-slot alignment reaches 92.0 PDMS, yielding a 0.5-point improvement. This result suggests that structured future-slot supervision provides additional benefits over generic dense feature alignment.

\begin{table}[t]
\centering
\caption{Comparison of dense future-feature alignment and future-slot alignment, both without action alignment.}
\label{tab:dense_future_alignment}
\small
\begin{tabular}{lc}
\hline
Future Alignment Target & PDMS $\uparrow$ \\
\hline
Dense future DINO features & 91.5 \\
Future slots & \textbf{92.0} \\
\hline
\end{tabular}
\end{table}

Effect of scorer supervision.
To investigate whether the benefits of world-model transfer depend on oracle-score supervision, we compare trajectory-only and full INTERACT variants under two scorer supervision strategies: oracle-score supervision and imitation-based winner-take-all (WTA) confidence supervision. We quantify the contribution of world-model transfer as the PDMS improvement over the corresponding trajectory-only baseline. As shown in Table~\ref{tab:scorer_supervision}, world-model transfer yields improvements of 1.3 and 1.4 PDMS under oracle-score and imitation-based WTA confidence supervision, respectively. The comparable gains indicate that the benefit of world-model transfer is retained without oracle-score supervision.

\begin{table}[t]
\centering
\caption{Effect of scorer supervision on the contribution of world-model transfer.}
\label{tab:scorer_supervision}
\small
\resizebox{\linewidth}{!}{
\begin{tabular}{lcc}
\hline
Scorer Supervision
& Trajectory-only
& Full INTERACT \\
\hline
Oracle-score
& 90.9 & 92.2 \\
Imitation-based WTA confidence
& 90.2 & 91.6  \\
\hline
\end{tabular}}
\end{table}

Effect of trajectory mode count.
To isolate the contribution of world-model supervision from the effect of increasing the number of trajectory hypotheses, we compare trajectory-only and INTERACT planners under both 16-mode and 64-mode configurations. As reported in Table~\ref{tab:64mode_comparison}, world-model supervision improves PDMS from 90.9 to 92.2 with 16 modes and from 92.4 to 93.8 with 64 modes. These results demonstrate consistent planning gains from world-model supervision beyond increasing the number of trajectory hypotheses.

\begin{table}[t]
\centering
\caption{Comparison of trajectory-only planning and INTERACT under 16-mode and 64-mode configurations.}
\label{tab:64mode_comparison}
\small
\resizebox{\linewidth}{!}{
\begin{tabular}{lccc}
\hline
Planner
& Number of Modes
& World-Model Supervision
& PDMS $\uparrow$ \\
\hline
Trajectory-only planner
& 16 & \xmark & 90.9 \\
INTERACT
& 16 & \cmark & 92.2 \\
Trajectory-only planner
& 64 & \xmark & 92.4 \\
INTERACT
& 64 & \cmark & \textbf{93.5} \\
\hline
\end{tabular}}
\end{table}